\pdfoutput=1

\documentclass[11pt]{article}

\usepackage[final]{coling}

\usepackage{times}
\usepackage{latexsym}

\usepackage[T1]{fontenc}

\usepackage[utf8]{inputenc}

\usepackage{microtype}
\usepackage{url}

\usepackage{inconsolata}

\usepackage{graphicx}
\usepackage{multirow}
\usepackage{booktabs}
\usepackage[normalem]{ulem}
\useunder{\uline}{\ul}{}

\title{Solid-SQL: Enhanced Schema-linking based In-context Learning for Robust Text-to-SQL}

\author{
 \textbf{Geling Liu\textsuperscript{1}\thanks{Corresponding author}\thanks{Key Laboratory of Aerospace Information Security and Trusted Computing, Ministry of Education}},
 \textbf{Yunzhi Tan\textsuperscript{2}},
 \textbf{Ruichao Zhong\textsuperscript{2}},
 \textbf{Yuanzhen Xie\textsuperscript{2}},
\\
 \textbf{Lingchen Zhao\textsuperscript{1}\footnotemark[2]},
 \textbf{Qian Wang\textsuperscript{1}\footnotemark[2]},
 \textbf{Bo Hu\textsuperscript{2}\footnotemark[1]},
 \textbf{Zang Li \textsuperscript{2}},
\\
 \textsuperscript{1}School of Cyber Science and Engineering, Wuhan University, China
\\
 \textsuperscript{2}Big Data and AI Platform Department, Tencent, China
\\
 \{liugl-, lczhaocs, qianwang\}@whu.edu.cn\\
 \{boristan, answerzhong, ashexie, harryyfhu, gavinzli\}@tencent.com
}

\begin{document}
\maketitle
\begin{abstract}


Recently, large language models (LLMs) have significantly improved the performance of text-to-SQL systems. Nevertheless, many state-of-the-art (SOTA) approaches have overlooked the critical aspect of system robustness. Our experiments reveal that while LLM-driven methods excel on standard datasets, their accuracy is notably compromised when faced with adversarial perturbations. To address this challenge, we propose a robust text-to-SQL solution, called Solid-SQL, designed to integrate with various LLMs. We focus on the pre-processing stage, training a robust schema-linking model enhanced by LLM-based data augmentation. Additionally, we design a two-round, structural similarity-based example retrieval strategy for in-context learning. Our method achieves SOTA SQL execution accuracy levels of 82.1\% and 58.9\% on the general Spider and Bird benchmarks, respectively. Furthermore, experimental results show that Solid-SQL delivers an average improvement of 11.6\% compared to baselines on the perturbed Spider-Syn, Spider-Realistic, and Dr. Spider benchmarks.
\end{abstract}
\section{Introduction}

\begin{figure}[t]
\centering
\includegraphics[width=0.95\columnwidth]{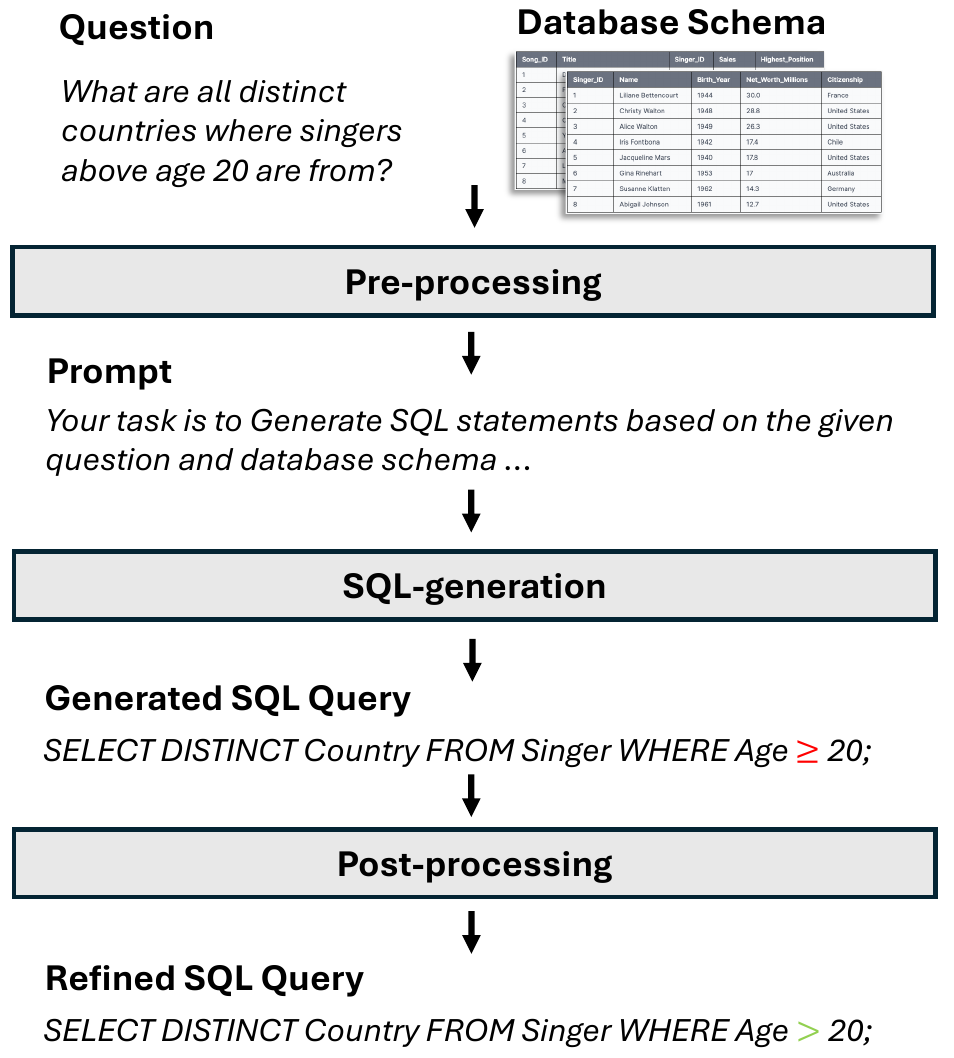}
\caption{The general three-stage pipeline of LLM-based text-to-SQL systems.}
\label{fig1}
\end{figure}

Text-to-SQL serves as an automated tool that facilitates the transformation of natural language into structured query language (SQL) commands, enabling individuals without specialized knowledge and skills to write SQL and query databases~\citep{baig2022natural}. 
Traditional text-to-SQL techniques have relied on rigid syntax tree templates~\citep{xu2017sqlnet, guo2019towards, wang2020rat} or supervised fine-tuning of sequence-to-sequence models~\citep{xie2022unifiedskg, scholak2021picard} for executing the transition from text to SQL. However, the recent past has witnessed a surge in the application of LLMs in text-to-SQL operations, proving their efficacy~\citep{gao2024text, pourreza2023din}. State-of-the-art approaches that top text-to-SQL leaderboards, such as Spider \citep{yu2018spider} and BIRD \citep{li2023can}, leverage advanced Large Language Models (LLMs) like GPT-4 \citep{achiam2023gpt} for SQL generation.

Considering the role of text-to-SQL in sensitive domains such as finance and healthcare, where system reliability and security are of critical importance, the robustness of text-to-SQL systems is essential, which, however, has not received adequate consideration in LLM-based text-to-SQL systems.
A robust text-to-SQL system should have the ability to maintain the correct SQL output when faced with adversarial perturbations in the text or database~\citep{pi2022towards}, such as changes in sentence structure, synonym descriptions, etc.
Experimental results reveal that leading LLM-based methods~\citep{dong2023c3, wang2024mac, li2024dawn} perform poorly on benchmarks that aim at testing the text-to-SQL robustnes, like Spider-Syn~\citep{gan2021towards}, Spider-Realistic~\citep{deng2021structure} and Dr.Spider~\citep{changdr}. 
However, efforts to enhance text-to-SQL robustness~\citep{furst2024evaluating, shen2023sequential, zhuo2023robustness} continue to be centered around traditional sequence-to-sequence architectures, with limited exploration of LLM-based alternatives~\citep{zhuo2023robustness}.

As illustrated in Figure~\ref{fig1}, an LLM-based text-to-SQL system comprises three distinct stages: pre-processing, SQL-generation, and post-processing. These stages are tasked with parsing the input question to synthesize effective prompts, querying the LLM to produce SQL statements, and refining the generated SQL, respectively. In the SQL-generation stage, to achieve good results, the employed LLMs typically have a very large size or are closed-source, making them difficult to fine-tune for rubustness. The post-processing phase entails refining the already generated SQL, with these refinements being independent of disturbances on the input side. The pre-processing stage, in contrast, deals with disruptions originating from the textual and tabular inputs. Therefore, handling in the pre-processing stage is crucial for enhancing the robustness of the LLM-based text-to-SQL systems. Specifically, how to process the text and schema to obtain a prompt that can stabilize performance in the SQL-generation stage is the problem we need to solve.

In this paper, to address the aforementioned issues, we design a robust pre-processing pipeline, called Solid-SQL, to generate prompts for SQL-generation. We craft the necessary pre-processing steps based on the components required for the prompt. A SQL statement consists of two components: first, the syntactic framework that determines the structure and logic of the statement; and second, the database schema, which includes the specific names of the tables and columns being accessed. To guide from both aspects, we aim to include pre-selected schemas and SQL statement examples with similar structures within our prompt. For robust schema selection, we specifically utilize LLMs to generate varied data for adversarial training and format training data specially for schema linking tasks to fine-tune a language model, addressing the lack of relevant datasets. To assist with in-context learning, we design effective methods for extracting text and SQL skeletons based on the chosen schemas and retrieve relevant SQL statement examples based on the similarity of these skeletons. When constructing the prompt, we incorporate explicit attention mechanisms to stabilize the output for inputs that have been perturbed.

Our contributions are summarized as follows: 
\begin{itemize}
    \item We address the existing gap in discussions on the robustness of LLM-based text-to-SQL systems. 
    To address this, we propose \textbf{Solid-SQL}, a pre-processing pipeline designed to enhance the robustness of LLM-based text-to-SQL systems in generating SQL.

    \item We design several effective modules, including a robust schema-linking model, example retrieval methods, and an explicit attention mechanism.
    Moreover, we validate Solid-SQL's universality and applicability through its integration with various SQL-generation LLMs.

    \item We conduct extensive experiments, demonstrating that Solid-SQL achieves SOTA performance on general benchmarks and significantly outperforms existing solutions on robustness benchmarks.
    Additionally, the effectiveness of the modules is validated through ablation studies.

\end{itemize}

\section{Related Work}

\subsection{LLM-based Text-to-SQL Techniques}
In tandem with the rapid advancement and widespread adoption of large language models (LLMs) across various natural language processing (NLP) domains, the text-to-SQL field has also seen significant benefits from recent methodological breakthroughs involving LLMs, achieving notable performance milestones. LLMs demonstrate impressive zero-shot reasoning and domain generalization capabilities, contributing to unprecedented achievements on the cross-domain Spider leaderboard~\citep{yu2018spider}. For instance, C3~\citep{dong2023c3} is a zero-shot text-to-SQL methodology that enhances ChatGPT through three designs: Clear Prompting for effective input; Calibration with Hints to correct model biases; and Consistent Output to ensure query reliability. The Chain-of-Thought approach~\citep{wei2022chain} has also been applied to text-to-SQL tasks. DIN-SQL~\citep{pourreza2023din} tackles the text-to-SQL task by decomposing it into four modules: schema linking, query classification and decomposition, SQL generation, and self-correction, each implemented using prompting techniques to leverage the granular capabilities of LLMs. Some methods further explore LLM's ability of in-context learning. DAIL-SQL~\citep{gao2024text} has revitalized the SOTA on Spider through a comprehensive examination of in-context learning, investigating the optimal selection of examples and their proper organization in prompts within a few-shot scenario. Other research has explored the selection of few-shot demonstrations by synthesizing in-domain examples~\citep{chang2023selective} and retrieving question skeletons~\citep{guo2023prompting}. Furthermore, MAC-SQL~\citep{wang2024mac} and CHESS~\citep{talaei2024chess} employ multi-agent collaboration for Text-to-SQL tasks. In addition to maximizing the ability to explore the LLMs without modifying it, other options involve fine-tuning the model. Approaches in DAIL-SQL~\citep{gao2024text}, DTS-SQL\citep{pourreza2024dts}, and CodeS~\citep{li2024codes} aim to enhance the capabilities of open-source LLMs through supervised fine-tuning, striving to compete with or surpass their larger, proprietary counterparts. 

Although these methods have achieved impressive results on the leaderboards, only a few of them have been evaluated for robustness~\citep{li2024codes, gao2024text}. Moreover, according to our experiments, the performance of many in-context learning based methods on robustness benchmarks appears to be somewhat inferior compared to their performance on the Spider and BIRD~\citep{li2023can} leaderboards.

\subsection{Adversarial Robustness}
Despite the remarkable performance of neural networks across various domains, they continue to exhibit significant vulnerabilities when subjected to perturbations~\citep{szegedy2014intriguing}.
This susceptibility is not only evident in traditional neural networks but has also been observed in systems such as text-to-SQL models~\citep{shen2023sequential}, where adversarial inputs can lead to degraded performance.
LLMs show potential as zero-shot text-to-SQL parsers, but their performance declines when faced with adversarial attacks and domain generalization disturbances, exhibiting varying levels of robustness in response to different types of perturbations~\citep{zhang2023towards}. It has been substantiated that removing explicitly stated column names~\citep{deng2021structure} or replacing database schema-related content with synonyms~\citep{gan2021towards} in the question will compromise the accessibility and accuracy of the generated SQL. Besides, confusion on the table side (e.g., substituting column descriptions or incorporating distracting columns within the table) will further undermine the precision of text-to-SQL systems~\citep{pi2022towards}. And for a holistic robustness assessment, Dr. Spider~\citep{changdr}, a diagnostic benchmark encompassing 15,000 perturbed examples that cover a multitude of perturbation types from three perspectives: the database, natural language questions, and SQL, has been unveiled. 

To enhance the robustness of text-to-SQL systems, several strategies have been employed, including manually adding synonym annotations to the schema to provide a more precise description~\citep{gan2021towards}, generating adversarial examples for adversarial training of the sequence-to-sequence model~\citep{pi2022towards, gan2021towards}, designing specialized training frameworks~\citep{deng2021structure} and crafting innovative encoding strategies to transition from text to SQL~\citep{shen2023sequential}.

However, these methods either require significant manual effort, are not suitable for new domains and large databases, or can just be applied to traditional encoder-decoder frameworks but not on the currently popular LLMs with large-scale parameters. In contrast, our robustness strategy is compatible with LLMs and ensures that text-to-SQL systems utilizing this strategy perform no worse than SOTA methods on conventional benchmarks, while markedly enhancing performance on robustness evaluation benchmarks.
\section{Methodology}

\begin{figure*}[t]
\centering{
\includegraphics[width=\textwidth]{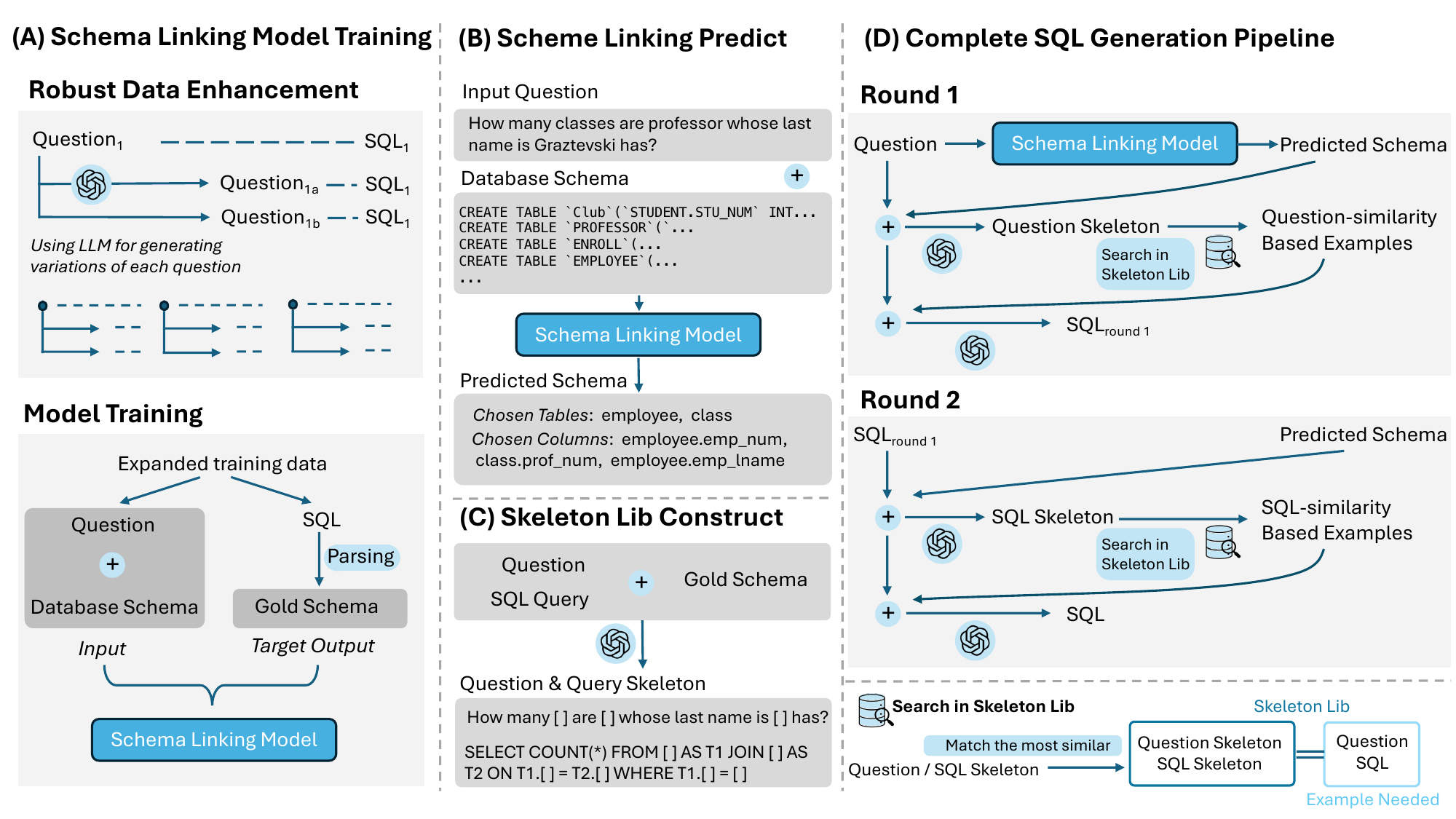}
\caption{The pipeline of Solid-SQL.}
\label{fig2}}
\end{figure*}

\subsection{Problem Definition}
Text-to-SQL is a task that generates a SQL statement for querying a database based on a natural language text which is a demand for some information about the database. It can be represented as $S = M(Q, SC)$, where $S$ is the generated SQL statement, $M$ is the text-to-SQL system, $Q$ is the natural language text, i.e., the question, and $SC$ is the database schema.

Our robustness goal is to stabilize the output of the text-to-SQL system when faced with perturbations. Specifically, without affecting the fundamental goal of $Q$, adding perturbations to $Q$ yields $Q^*$. A robust LLM-based text-to-SQL strategy satisfies $DB(M(Q, SC)) = DB(M(Q^*, SC))$, where $DB$ is the access database.

\subsection{Overview}

Solid-SQL is a novel plug-in solution for robust text-to-SQL, which uses a carefully designed pre-processing pipeline to extract the elements required for prompt composition. It includes a robust scheme linking model,  an effective example retrieve method and an explicit attention mechanism.

Figure~\ref{fig2} shows the pipeline of Solid-SQL. Firstly, we employ an LLM to introduce minor perturbations to the text of text-to-SQL training data, while preserving its semantic integrity. In this way, we create an augmented dataset of clean and perturbed data. This augmented dataset is then used to fine-tune a language model for schema linking(Figure~\ref{fig2}(A)). Subsequently, leveraging the outcomes of schema linking(Figure~\ref{fig2}(B)), we ask an LLM to extract and remove domain-specific information and value information from the input text query, to derive the query's skeleton(Figure~\ref{fig2}(C)). This skeleton is matched against a pool of candidate skeletons to retrieve an appropriate number of relevant samples as examples based on similarity. The question, complete schema, and examples are combined into a prompt, with explicit emphasis on the filtered schema, to query the LLM for SQL generation in the first round(Figure~\ref{fig2}(D)-Round 1). Following this, the SQL generated from the first round is parsed to extract its backbone, and additional examples are retrieved for the second round of SQL generation, culminating in the final SQL output (Figure~\ref{fig2}(D)-Round 2).

\subsection{Schema Linking}
The schema linking task is a preliminary step of text-to-SQL, simplifying the generation of SQL queries. Its goal is to select the actual tables and columns to be accessed from the entire database schema based on a given question. For schema linking, since the input contains numerous SQL statements that define the database structure, it is challenging for a base LLM with general capabilities to understand such input and produce output in the expected format. Therefore, we fine-tune a model to complete the task.

\subsubsection{Robust Data Enhancement}
In order to improve the robustness of our schema linking model, we first expanded the original text-to-SQL training dataset. The data in the training set is in the form of a triplet ${Q, SC, S}$, where $Q$ is the input text query, $SC$ is the complete database schema, and $S$ is the correct SQL query output. We use an LLM to rewrite $Q$, including changing the sentence structure and replacing synonyms (i.e., substituting 'singer' with 'musician'), resulting in new questions $Q_1$ and $Q_2$. Then, we add the new triplets ${Q_1, SC, S}$ and ${Q_2, SC, S}$ to the training dataset. This expanded training set introduces perturbations and adversarial examples, and the model trained with this augmented dataset can effectively improve its robustness.

\subsubsection{Model Training}
We set the target operation to choose schema in Solid-SQL, as shown in Figure~\ref{fig2}(B), formulated as $\{T, C\} = G(Q, SC)$, where $T = \{T_{1}, T_{2}, \dots, T_{|T|}\}$ and $C = \{C_{1}, C_{2}, \dots, C_{|C|}\}$ are the selected tables and columns, respectively, and $G$ is a fine-tuned generation model for schema linking. The absence of a training set for schema linking is problematic, and existing text-to-SQL training datasets offer data in the form of $(Q, SC)-S$ pairs without providing the chosen schema. Therefore, we parse the SQL statement $S$ related to question $Q$ to obtain the ground truth ${T, C}$. 


\subsection{Relative Example Retrieval}
In crafting prompts for LLMs, the selection of task-relevant examples is paramount, as it enhances the model's comprehension and task performance by leveraging contextual adaptability, knowledge transfer, and ambiguity mitigation. 

Verified by DAIL-SQL \citep{gao2024text}, examples formatted as pairs consisting of a text query and the corresponding correct SQL statement benefit the in-context learning for the text-to-SQL task, rather than containing only the text or the SQL query. We define the example set as $E = \{E_1, E_2, \ldots, E_N\}$, where each $E_i$ corresponds to a question-SQL pair, denoted as $(Q_i, S_i)$.

\subsubsection{Question Skeleton Similarity-Based}
The correlation between two SQL statements, $S_i$ and $S_j$, suggests a corresponding relationship between their associated questions, $Q_i$ and $Q_j$. To guide the LLM towards generating the desired SQL, it is reasonable to select analogous questions from the candidate set as examples, based on the target question $Q$.

However, the emphasis on similarity should focus on the structural alignment of the SQL statements rather than their thematic proximity. To achieve this, it is crucial to abstract the target question $Q$ by removing domain-specific details and value-related content, revealing its core structure, or 'skeleton', denoted as $Q^{\star}$. $Q^{\star}$ serves as the foundation for identifying analogous questions within the candidate set, with a focus on those exhibiting a similar structural pattern. By aligning the examples with $Q^{\star}$, the model can more accurately identify the underlying patterns and relationships essential for converting natural language queries into executable SQL commands. This method ensures that the LLM's in-context learning is attuned to the structural intricacies critical for the task.

To derive $Q^{\star}$ from $Q$, we leverage the language understanding capabilities of an LLM and employ a prompting-based technique. We input the question $Q$ along with the schema inferred by our schema linking model into a universal LLM and parse the output to extract $Q^{\star}$. As shown in Figure~\ref{fig2}(C), this process obscures the domain-specific information and values, leaving only the question's skeletal structure. This extraction process should be applied to both the questions $Q_i$ within the candidate library ${(Q_1, S_1), (Q_2, S_2), \ldots, (Q_{|E|}, S_{|E|})}$ and the given target question $Q$ itself. Based on the cosine similarity between $Q^{\star}$ and the set ${{Q^{\star}_1}, {Q^{\star}2}, \ldots, {Q^{\star}{|E|}}}$, we can identify the top $N$ most similar candidate skeletons, corresponding to example pairs ${(Q_1, S_1), (Q_2, S_2), \ldots, (Q_N, S_N)}$.

\subsubsection{SQL Skeleton Similarity-Based}

In addition to indirectly utilizing questions to match the samples to be retrieved, direct matching can also be achieved through the similarity of SQL statements. As shown in Figure~\ref{fig2}(D), after the SQL generation is completed by the LLM in round 1, we can select examples for the SQL generation in round 2 based on the similarity of the SQL skeletons.

We extract the skeleton $S^{\star}$ of an SQL statement $S$ by identifying and manipulating its various components using an SQL parsing tool\footnote{\url{https://github.com/tobymao/sqlglot} }. This process involves parsing the SQL statement to generate a syntax tree, then recognizing and replacing the table names, column names, and values within it, while preserving the SQL keywords and logical structure. Ultimately, a skeleton that contains only placeholders and the structure of the SQL is produced. This extraction process is applied to both the candidate library and the generated SQL from round 1.

We employ the edit distance derived from the parse tree of an SQL skeleton to quantify the structural similarity, which provides an analytical approach that emphasizes the logical framework of SQL statements rather than their superficial textual similarities. This technique enables a more precise identification of key element correspondences than calculating cosine similarity of embedded vectors~\citep{gao2024text}. It can also effectively discounts disparities arising from diverse expressive forms or functional applications. Consequently, it facilitates a more robust assessment of the conceptual similitude within SQL statements.

\subsection{Information Utilization}
We design suitable prompt templates to integrate existing information and query LLMs accomplish SQL-generation effectively. A key consideration is the use of an explicit attention mechanism to embed table and column names filtered through schema linking into the prompt. Existing approaches~\citep{gao2024text} often present only the filtered schema information to the LLM, reducing token count and excluding unnecessary information. However, this method has a significant drawback: if crucial schema information is omitted in the last step, the LLM will be unable to generate the correct SQL statement. In contrast, we use "focus on" to emphasize schema elements that are more likely to form the final SQL statement to the LLM. Thus the model still has a comprehensive view of the entire schema information while understanding the priority, ensuring the stability and fault tolerance of the LLM in generating SQL statements when faced with perturbation.

\section{Experiments}
\begin{table*}[ht]
\renewcommand{\arraystretch}{1.3}
\caption{The EX (Excute Accuracy) and EM (Exact Match) of Solid-SQL on Spider dev set, Spider-Syn test set and Spider-Realistic test set, comparing with SOTA open source prompting based methods.}
\label{table1}
\centering
\resizebox{\linewidth}{!}{
\begin{tabular}{cl|cc|cc|cc|cc}
\toprule
\multirow{2}{*}{Alternative LLM} & \multicolumn{1}{c|}{\multirow{2}{*}{Method}} & \multicolumn{2}{c|}{Spider}       & \multicolumn{2}{c|}{Spider-Syn}       & \multicolumn{2}{c|}{Spider-Realistic} & \multicolumn{2}{c}{Avg}              \\ \cmidrule(l){3-10} 
                                    &                         & EX                & EM                & EX                & EM                & EX                & EM                & EX                & EM               \\ \midrule
\multirow{5}{*}{Llama3-70b}         & DAIL-SQL~\citep{gao2024text}   & 60.5              & 42.4              & 58.0              & 33.3              & 59.8              & 41.5              & 59.4              & 39.1             \\
                                    & MAC-SQL~\citep{wang2024mac}      & 76.3              & 27.9              & 65.1              & 21.7              & 71.7              & 24.4              & 71.0              & 24.7             \\
                                    & DIN-SQL~\citep{pourreza2023din}    & \multicolumn{2}{c|}{\textbackslash{}} & \multicolumn{2}{c|}{\textbackslash{}} & \multicolumn{2}{c|}{\textbackslash{}} & \multicolumn{2}{c}{\textbackslash{}} \\ 
                                    & Solid-SQL (round1)           & 81.5              & 59.3              & 74.2              & 52.3              & 76.6              & 52.0              & 77.4              & 54.5             \\
                                    & Solid-SQL (round2)           & 82.1              & 61.0              & 74.6              & 54.7              & 77.1              & 54.3              & 77.9              & 56.7             \\ \midrule
\multirow{5}{*}{deepseek-13b}       & DAIL-SQL~\citep{gao2024text}       & 68.6              & 52.9              & 50.1              & 37.4              & 60.2              & 48.0              & 59.6              & 46.1             \\
                                    & MAC-SQL~\citep{wang2024mac}    & 50.0              & 5.8               & 39.7              & 3.7               & 37.4              & 3.9               & 42.4              & 4.5              \\
                                    & DIN-SQL~\citep{pourreza2023din}    & \multicolumn{2}{c|}{\textbackslash{}} & \multicolumn{2}{c|}{\textbackslash{}} & \multicolumn{2}{c|}{\textbackslash{}} & \multicolumn{2}{c}{\textbackslash{}} \\
                                    & Solid-SQL (round1)       & 77.3              & 52.5              & 68.4              & 43.4              & 71.4              & 50.8              & 72.3              & 48.9             \\
                                    & Solid-SQL (round2)       & 77.8              & 51.6              & 68.5              & 43.4              & 72.3              & 50.4              & 72.9              & 48.5             \\ \midrule
\multirow{5}{*}{GPT-4o-mini}        & DAIL-SQL~\citep{gao2024text}    & 77.4              & 51.4              & 66.1              & 37.8              & 70.5              & 49.8              & 71.3              & 46.3             \\
                                    & MAC-SQL~\citep{wang2024mac}   & 78.1              & 37.8              & 68.7              & 29.4              & 76.0              & 39.0              & 74.3              & 35.4             \\
                                    & DIN-SQL~\citep{pourreza2023din}     & 70.3              & 36.4              & 64.3              & 33.4              & 62.0              & 38.8              & 65.5              & 36.2             \\
                                    & Solid-SQL (round1)      & 80.4              & 60.0              & 74.2              & 52.3              & 76.1              & 54.7              & 76.9              & 55.7             \\
                                    & Solid-SQL (round2)     & 79.9              & 60.3              & 74.6              & 53.3              & 76.7              & 54.3              & 77.1              & 56.0             \\ \bottomrule
\end{tabular}
}
\end{table*}

\subsection{Experimental Setup}

\subsubsection{Datasets}
We evaluate the performance of Solid-SQL on a simple clean test set called Spider, a difficult clean test set called Bird, and three perturbed test sets.

\textbf{Spider}~\citep{yu2018spider} is a dataset for semantic parsing and text-to-SQL, created by Yale students. It contains 10,181 questions and 5,693 SQL queries across 200 databases and 138 domains.

\textbf{Bird}~\citep{li2023can} is a large-scale text-to-SQL benchmark developed by Alibaba DAMO Academy. It features 12,751 question-SQL pairs, 95 databases, and spans over 37 domains.

\textbf{Spider-Syn}~\citep{gan2021towards} is an adapted version of the Spider dataset with 5,672 questions modified by replacing words with their synonyms, using 273 synonyms and 189 phrases. On average, there is one alteration per question.

\textbf{Spider-Realistic}~\citep{deng2021structure} is a perturbed evaluation set based on Spider. It has been manually adjusted to remove explicit column names while keeping SQL queries unchanged.

\textbf{Dr. Spider}~\citep{changdr} is a robustness evaluation benchmark for text-to-SQL models. It applies perturbations to the database, natural language query, and SQL components, and contains 15,000 pre- and post-perturbation examples.

\subsubsection{Evaluation Metrics}
We assess the performance of the text-to-SQL model by evaluating the quality of the generated SQL. Execution Accuracy (EX) is defined as the proportion of questions in the evaluation set for which the execution outcomes of both the predicted SQL queries and the ground-truth queries are the same. It is calculated relative to the total number of queries. The EX of the generated SQLs indicates how well the model meets the availability and precision requirements in real-world scenarios. We also use Exact Match Accuracy (EM) as an adjunct. EM is the portion of generated SQLs that totally match the ground truth SQL statements.

\subsubsection{LLMs}
Solid-SQL employs a prompting methodology that supports the use of various interchangeable LLMs. To validate the compatibility and generalizability of our proposed solution, we conducted experiments using four distinct LLMs for SQL generation. These models included both open-source and closed-source options.




\textbf{LLama3-70b}: An open-source LLM with 70 billion parameters by Meta AI, optimized for diverse NLP tasks including text generation and translation.

\textbf{Deepseek-coder-33b-instruct}: A 33-billion-parameter model from the Deepseek Coder series, leading in open-source code generation across multiple programming languages.

\textbf{GPT-4o-mini}: A compact version of GPT-4o, retaining core text capabilities with faster inference due to fewer parameters.

\textbf{GPT-4}: OpenAI's advanced generative pre-trained transformer, adept at complex tasks like essay writing and coding with high accuracy and creativity.

\subsubsection{Baselines}
We compare with other prompting-based text-to-SQL solutions which have SOTA performances on Spider and Bird.




\textbf{DAIL-SQL}~\citep{gao2024text}: Ranked second on the Spider leaderboard, DAIL-SQL employs a variety of example selection methods and a structured format for example organization. Leveraging GPT-4, it achieves high performance in SQL generation quality and query efficiency.

\textbf{MAC-SQL}~\citep{wang2024mac}: This method features a core decomposer agent for Text-to-SQL with few-shot chain-of-thought reasoning, supported by two auxiliary agents for obtaining subdatabases and refining SQL queries. The agents work in tandem, with the flexibility to integrate new tools or features for enhanced Text-to-SQL parsing.

\textbf{DIN-SQL}~\citep{pourreza2023din}: DIN-SQL breaks down the text-to-SQL task into sub-problems: schema linking, query classification \& decomposition, SQL generation, and self-correction. Utilizing prompting techniques, it demonstrates that LLMs can effectively solve these sub-problems when appropriately decomposed.

\textbf{CodeS}~\citep{li2024codes}: CodeS is an open-source language model series designed for text-to-SQL, offering high accuracy with smaller parameters compared to closed-source LLMs. It employs an incremental pre-training strategy on a SQL-specific corpus and addresses schema linking and domain adaptation challenges. Evaluations show CodeS achieves state-of-the-art performance on multiple text-to-SQL benchmarks.

\subsubsection{Solid-SQL Details}
In the deployment of Solid-SQL, we employ the LLama3-8B-Instruct model as the foundational architecture for the schema linking task. The model is subjected to a training regimen consisting of five epochs on an augmented dataset, which comprises approximately 22,000 question-SQL pair instances. For the extraction of question skeletons, we use the LLM align with SQL-generation. Furthermore, when assessing the cosine similarity between two question embeddings, we employ the bge-large-en-v1.5 embedding model before the computation.

\begin{table*}[t]
\caption{The EX (Execute Accuracy) of Solid-SQL (ours) and baselines on Dr.Spider. "Pert Level" is where the perturbation is added and "Pert Type" is the how the perturbation added.}
\label{table2}
\centering
\resizebox{\linewidth}{!}{
\begin{tabular}{cl|c|cc|cc}
\toprule
\multirow{2}{*}{Pert Level} & \multicolumn{1}{c|}{\multirow{2}{*}{Pert Type}} & \multirow{1}{*}{CodeS-15B} & \multicolumn{2}{c|}{Llama3-70b}                 & \multicolumn{2}{c}{GPT-4o-mini}                 \\ \cmidrule(l){4-7} 
                            & \multicolumn{1}{c|}{}                           &  \multicolumn{1}{c|}{\citep{li2023li}}              & \multicolumn{1}{c|}{MAC-SQL} & ours-round1 & \multicolumn{1}{c|}{MAC-SQL} & ours-round1 \\ \midrule
DB                          & DBcontent-equivalence                           & 47.6                           & \multicolumn{1}{c|}{58.4}    & {\ul 62.8}       & \multicolumn{1}{c|}{59.9}    & \textbf{63.2}    \\
                            & schema-abbreviation                             & \textbf{78.7}                  & \multicolumn{1}{c|}{72.9}    & 77.9             & \multicolumn{1}{c|}{74.1}    & {\ul 78.0}       \\
                            & schema-synonym                                  & 66.9                           & \multicolumn{1}{c|}{65.9}    & \textbf{72.8}    & \multicolumn{1}{c|}{66.8}    & {\ul 72.6}       \\ \midrule
NLQ                         & column-attribute                                & {\ul 68.9}                     & \multicolumn{1}{c|}{63.9}    & \textbf{69.2}    & \multicolumn{1}{c|}{65.7}    & {\ul 68.9}       \\
                            & column-carrier                                  & {\ul 79.1}                     & \multicolumn{1}{c|}{66.3}    & 78.9             & \multicolumn{1}{c|}{68.9}    & \textbf{79.8}    \\
                            & column-synonym                                  & 64.7                           & \multicolumn{1}{c|}{53.3}    & \textbf{70.3}    & \multicolumn{1}{c|}{54.8}    & {\ul 70.1}       \\
                            & column-value                                    & 76.3                           & \multicolumn{1}{c|}{68.8}    & \textbf{78.8}    & \multicolumn{1}{c|}{69.6}    & {\ul 77.4}       \\
                            & keyword-carrier                                 & \textbf{91.7}                  & \multicolumn{1}{c|}{89.5}    & 91.1             & \multicolumn{1}{c|}{90.5}    & {\ul 91.2}       \\
                            & keyword-synonym                                 & {\ul \textit{73.5}}            & \multicolumn{1}{c|}{62.0}    & \textbf{74.8}    & \multicolumn{1}{c|}{64.6}    & 73.1             \\
                            & multitype                                       & 69.4                           & \multicolumn{1}{c|}{58.3}    & \textbf{71.5}    & \multicolumn{1}{c|}{61.2}    & {\ul 70.9}       \\
                            & others                                          & \textbf{81.2}                  & \multicolumn{1}{c|}{69.8}    & 79.2             & \multicolumn{1}{c|}{71.1}    & {\ul 81.0}       \\
                            & value-synonym                                   & 71.9                           & \multicolumn{1}{c|}{60.1}    & {\ul 73.2}       & \multicolumn{1}{c|}{62.5}    & \textbf{73.9}    \\ \midrule
\multirow{5}{*}{SQL}        & comparison                                      & 71.9                           & \multicolumn{1}{c|}{75.3}    & {\ul 77.6}       & \multicolumn{1}{c|}{76.2}    & \textbf{77.8}    \\
                            & DB-number                                       & \textbf{85.9}                  & \multicolumn{1}{c|}{80.5}    & 83.7             & \multicolumn{1}{c|}{81.8}    & {\ul 84.6}       \\
                            & DB-text                                         & \textbf{80.7}                  & \multicolumn{1}{c|}{72.8}    & 79.6             & \multicolumn{1}{c|}{75.2}    & {\ul 80.2}       \\
                            & nonDB-number                                    & 84.0                           & \multicolumn{1}{c|}{84.0}    & {\ul 87.0}       & \multicolumn{1}{c|}{86.3}    & \textbf{89.7}    \\
                            & sort-order                                      & \textbf{84.9}                  & \multicolumn{1}{c|}{62.0}    & 77.3             & \multicolumn{1}{c|}{66.1}    & {\ul 79.5}       \\ \bottomrule
\end{tabular}
}
\end{table*}

\begin{table}[]
\caption{The EX (Execute Accuracy) on Bird dev set.}
\label{table3}
\centering
\begin{tabular}{lc}
\toprule
Method            & Bird \\ \midrule
GPT-4             & 46.2     \\
DIN-SQL + GPT-4   & 50.7     \\
DAIL-SQL + GPT-4  & 54.8     \\
MAC-SQL + GPT-4   & 50.6     \\
Solid-SQL + GPT-4 & 58.9     \\ \bottomrule
\end{tabular}
\end{table}

\subsection{Overall Performance}
We test the performance of Solid-SQL across various benchmarks and compared it with the baselines.

Table~\ref{table1} presents the performance of Solid-SQL on the Spider benchmark as well as its robustness test variants, Spider-syn and Spider-realistic. Leveraging the prompting-based nature of Solid-SQL and the compared baseline methods, we implemented a plugin-style approach to substitute various LLMs for SQL generation. Solid-SQL significantly outperforms the baselines in both execution accuracy (EX) and exact match (EM) across all datasets, with an average execution accuracy 12.4\% higher than the baselines and an average exact match that also exceeds the baseline. It is noteworthy that certain methods exhibit a strong dependency on specific LLMs, seeing that DAIL-SQL's performance decreases heavily on LLama compared with GPT series and DIN-SQL is even unable to output reasonable SQL statements when using Deepseek and Llama. In contrast, Solid-SQL performs well on all the test LLMs, showing versatility and compatibility.

Table~\ref{table2} demonstrates the execution accuracy of Solid-SQL when generating SQL queries under various levels and types of perturbations within the Dr.Spider dataset. Due to MAC-SQL's most stable performance in cooperation with different LLMs among the baselines in Table~\ref{table1}, we choose it as the object of comparison. The results clearly show that our Solid-SQL approach significantly outperforms MAC-SQL and matches the current state-of-the-art model in robustness, CodeS-15B, which has been fine-tuned with extensive data. Additionally, it is evident that Solid-SQL has a distinct advantage in terms of robustness against perturbations that involve the use of synonyms.

Table~\ref{table3} displays the experimental results on the Bird benchmark, which similarly demonstrates the consistent performance of Solid-SQL under complex requirements.

\subsection{Ablation Study \& Hyper-parameter Study}
\subsubsection{Schema Linking Training}
Table~\ref{table4} presents the results of an ablation study on schema linking training. Compared to the baseline model without supervised fine-tuning (SFT), the model fine-tuned with the basic training set shows a significant improvement of approximately 25\% in the accuracy of column name choosing, underscoring the necessity and efficacy of SFT. Moreover, employing a robustness-enhanced augmented training set with added perturbations further improves the accuracy of schema linking, especially on perturbed benchmarks such as Spider-Syn and Spider-Realistic, with an enhancement of about 2\%, highlighting the contribution to the robustness of schema linking.

\subsubsection{Explicit Attention in Prompt}
Table~\ref{table5} presents the results of an ablation study of the explicit attention mechanism. The study examines how the existence of it in prompts impacts SQL-generation. The study reveals that incorporating this mechanism significantly improves SQL exact match accuracy across three datasets, especially when dealing with synonym perturbations, as seen in Spider-Syn. Additionally, it is noteworthy that the positive effect of this design is more pronounced when the number of examples in the prompt is reduced. This suggests that in scenarios where conserving tokens is crucial, our design can effectively enhance the performance of LLMs in generating SQL queries.

\subsubsection{Number of Retrieved Examples}
Table~\ref{table6} presents a study on the optimal number of examples to include in prompts for in-context learning. Although performance across various benchmarks varies with different numbers of examples, there is a general trend where the ability of LLMs to generate SQL queries initially strengthens and then weakens as the number of examples increases. Based on the average performance, we ultimately set the number of examples, denoted as $N$, to be recalled to 7.



\begin{table}[]
\caption{The ablation study of SFT-based schema linking in Solid-SQL. Values in the table are the searching accuracy of cloumn names. "w/o SFT" referring to the base model without finetuning, "w/o Enhance" referring to being finetuned with basic training data, and "with Enhance" referring to being finetuned with enhanced training data.}
\label{table4}
\centering
\resizebox{\linewidth}{!}{
\begin{tabular}{clll}
\toprule
                      & \textbf{Spider}      & \textbf{Syn}  & \textbf{Realistic} \\ \midrule
w/o SFT      & 65.6                 & 59.9                 & 62.5                  \\ \midrule
w/o Enhance  & 88.6                 & 82.4                 & 83.1                      \\
with Enhance & 89.7 ($\uparrow$\textbf{1.1}) & 84.2 ($\uparrow$\textbf{1.9}) & 85.1 ($\uparrow$\textbf{2.0}) \\ \bottomrule
\end{tabular}
}
\end{table}

\begin{table}[] 
\caption{The ablation study of the design of explicit attention in prompt construction. Values in the table are the EX of SQL generated by one-round Solid-SQL with Llama3-70b. $N$ is the number of examples retrieved for in-context learning. "with focus" and "w/o focus" referring to have the design or not, respectively.}
\label{table5}
\centering
\resizebox{\linewidth}{!}{
\begin{tabular}{cclll}
\toprule
$N$                     &            & \textbf{Spider}     & \textbf{Syn}        & \textbf{Realistic}        \\ \midrule
\multirow{2}{*}{3}      & w/o focus  & 78.5                & 66.9                & 74.2                      \\
                        & with focus & 81.7 ($\uparrow$\textbf{3.2})&73.8 ($\uparrow$\textbf{6.9})&76.0 ($\uparrow$\textbf{1.8})\\ \midrule
\multirow{2}{*}{9}      & w/o focus  & 79.8                & 67.5                & 74.8                      \\
                        & with focus & 81.1 ($\uparrow$\textbf{1.3})&73.9 ($\uparrow$\textbf{6.4})&76.4 ($\uparrow$\textbf{1.6})\\ \bottomrule
\end{tabular}
}
\end{table}

\begin{table}[]
\caption{The study of how many examples should be added into the prompt. $N$ is the number of examples retrieved ranging from 1 to 9. Values in the table are the executing accuracy of SQL generated in round 1 of Solid-SQL.}
\label{table6}
\centering
\resizebox{\linewidth}{!}{
\begin{tabular}{lccccc}
\toprule
\multicolumn{1}{c}{$N$}                & 1     & 3     & 5     & 7              & 9     \\ \midrule
\textbf{Spider}            & 80.9  & 81.7  & 81.2  & 81.5           & 81.1  \\
\textbf{Syn}        & 74.0  & 73.8  & 73.7  & 74.2           & 73.9  \\
\textbf{Realistic} & 75.6  & 76.0  & 76.4   & 76.6           & 76.4  \\ \midrule
\textbf{Avg}               & 76.83 & 77.17 & 77.10 & \textbf{77.43} & 77.13 \\ \bottomrule
\end{tabular}
}
\end{table}

\section{Limitations}

The Solid-SQL approach offers opportunities for enhancement, particularly in the procedural design. We could define conditions for advancing to a second round of queries, which would streamline the process by eliminating unnecessary steps and increase algorithmic efficiency. Furthermore, the plugin-based architecture of Solid-SQL suggests the possibility of integrating it with other methodologies to achieve performance improvements, an option we have not yet fully investigate due to time constraints.

\section{Conclusion}
In this paper, we present Solid-SQL, a robust text-to-SQL solution designed to address the robust limitations of current SOTA LLM-based methods.
By focusing on pre-processing techniques and integrating a robust schema-linking model, along with a two-round example retrieval strategy, Solid-SQL significantly improves SQL execution accuracy on both standard and adversarially perturbed benchmarks.
Solid-SQL achieves SOTA performance on general benchmarks and an average advancement of 11.6\% over baselines on perturbed benchmarks, proving its effectiveness in enhancing the robustness of text-to-SQL systems.
This work highlights the importance of system robustness in the development of text-to-SQL models and lays the groundwork for future research in this field.

\section*{Acknowledgments}
We thank the anonymous reviewers for their helpful and valuable feedback. This work was partially supported by NSFC under Grants U2441240, U21B2018, and 62302344.

\bibliography{custom}

\end{document}